\documentclass[letterpaper, 10pt, conference]{ieeeconf}      

\IEEEoverridecommandlockouts                              
\usepackage{graphicx} 
\usepackage{amsmath} 
\usepackage{amsfonts}
\usepackage{mathtools} 

\usepackage{multirow}
\usepackage{tabularx}
\usepackage{caption}
\usepackage{subcaption}
\usepackage{hyperref}
\usepackage{float}
\usepackage{eucal} 
\usepackage{siunitx}
\usepackage{todonotes}

\usepackage{algorithm} 
\usepackage[noend]{algpseudocode}

\DeclareMathOperator*{\argmin}{arg\,min}

\newcommand{\truesystemstate}[1]{\mathcal{G}_{#1}}
\newcommand{\estimatecost}[2]{E_{#1}^{#2}}

\newcommand{\estimatesystemstate}[1]{\bar{\mathcal{G}}_{#1}}

\newcommand{\robot}{\mathcal{R}}

 \newcommand{\truerelsystemstate}[1]{\mathcal{G}r_{#1}}
\newcommand{\relestimatecost}[2]{Er_{#1}^{#2}}
\newcommand{\estimaterelsystemstate}[1]{\bar{\mathcal{G}}r_{#1}}

\newcommand{\trans}[2]{^{#2}\bold{p}_{#1}}
\newcommand{\rot}[2]{^{#2}\bold{R}_{#1}}
\newcommand{\yaw}[2]{^{#2}\psi_{#1}}
\newcommand{\measure}[2]{^{#2}\bold{z}_{#1}}

\newcommand{\state}[1]{\bold{x}_{#1}}
\newcommand{\statecov}[1]{\bold{P}_{#1}}
\newcommand{\cinput}[1]{\bold{u}_{#1}}
\newcommand{\dotstate}{\dot{\bold{x}}}
\newcommand{\statenoise}[1]{\bold{\omega}_{#1}}

\newcommand{\pdot}[2]{^{#2}\dot{\bold{p}}_{#1}}
\newcommand{\yawdot}[2]{^{#2}\dot{{\psi}}_{#1}}
\newcommand{\vtrans}[2]{^{#2}\bold{v}_{#1}}
\newcommand{\vyaw}[2]{^{#2}\bold{\omega}_{#1}}
\newcommand{\gnoise}[2]{^{#2}\bold{\eta}_{\bold{#1}}}

\newcommand{\mnoise}[2]{^{#2}\bold{\epsilon}_{\bold{#1}}}

\newcommand{\pertubmatrix}[1]{\Tilde{\chi}_{#1}}

\newcommand{\defeq}{\vcentcolon=}
 
\newcommand{\detector}[1]{\boldsymbol{\phi}_{#1}}

\newcommand{\Bold}[2]{\bold{#1}_{#2}}
 
\title{\LARGE \bf Vision-based Multi-MAV Localization with Anonymous Relative Measurements Using Coupled Probabilistic Data Association Filter}

\author{Ty Nguyen$^{1*}$, Kartik Mohta$^{2*}$, Camillo J. Taylor$^{1}$, Vijay Kumar$^{1}$
\thanks{This work was supported by the MAST Collaborative Technology Alliance - Contract No. W911NF-08-2-0004, ARL grant W911NF-08-2-0004, ONR grants N00014-07-1-0829, N00014-14-1-0510, ARO grant W911NF-13-1-0350, NSF grants IIS-1426840, IIS-1138847, DARPA grants HR001151626, HR0011516850, and supported in part by the Semiconductor Research Corporation (SRC)
and DARPA.}
\thanks{$^{1}$ The authors are with the
GRASP Lab, University of Pennsylvania, Philadelphia, PA 19104 USA. {
        {\tt\footnotesize email: \{tynguyen, cjtaylor, kumar\}}@seas.upenn.edu}
}%
\thanks{$^{2}$ The author is with Autel Robotics, USA
}
\thanks{$^{*}$ Authors contribute similarly
}
}

\begin{document}

\maketitle
\thispagestyle{empty}
\pagestyle{empty}

\begin{abstract}
We address the localization of robots in a multi-MAV system where external infrastructure like GPS or motion capture system may not be available. We introduce a vision plus IMU system for localization that uses relative distance and bearing measurements. Our approach lends itself to implementation on platforms with several constraints on size, weight, and payload (SWaP). Particularly, our framework fuses the odometry with anonymous, visual-based robot-to-robot detection to estimate all robot poses in one common frame, addressing three main challenges: 1) initial configuration of the robot team is unknown, 2) data association between detection and robot targets is unknown, and 3) vision-based detection yields false negatives, false positives, inaccurate, noisy bearing and distance measurements of other robots. Our approach extends the Coupled Probabilistic Data Association Filter (CPDAF)~\cite{cpdaf} to cope with nonlinear measurements. We demonstrate the superior performance of our approach over a simple VIO-based method in a simulation using measurement models obtained from real data. We also show how on-board  sensing,  estimation and control can be used for formation flight.
 
\end{abstract}


\section{Introduction}
Multi-robot systems are of interest for their potential in performing tasks which may not be feasible or desirable to do with only a single robot in applications such as perimeter surveillance~\cite{vijayperimetersurveillance_1, davidperimetersurveillance}, patrolling missions~\cite{marino2013decentralized, portugal2016cooperative}, searching operations~\cite{marjovi2009multi, basilico2011exploration}, and formation control~\cite{franchiformationcontrol, vijayformationcontrol_1, vijayformationcontrol_2}. For example, the task of surveilling a large area is often infeasible for one robot due to the robot's limited coverage but can be accomplished by a team of robots under proper coordination. 
A major requirement for these applications is that the robots need to be localized within a common reference frame.
That way, each robot can execute its designated subtask correctly and the team can collaboratively complete the full task.
This requirement becomes trivial when there is a single global coordinate system can provide the state estimate for all robots, such as GPS, motion capture systems, aerial images~\cite{poulet2018self}. However, such systems are often not available or reliable.

Another solution to this problem is to launch the robots in a predetermined spatial configuration with a common frame and let robots localize within this frame. This solution is obviously time consuming and requires significant effort since we either need to displace the robots at predetermined poses or to measure the relative poses between the robots at the beginning.  


Alternative solutions rely on local sensing modalities such as bearing, range, and camera imaging to measure the instantaneous, relative pose between pairs of robots. These modalities can provide measurements which are either landmark features of the environment or direct relative poses. By allowing robots to collaboratively localize using these relative measurements, they can self-localize in a common frame. 


\begin{figure}[t]
    \centering
    \includegraphics[width=0.8\linewidth]{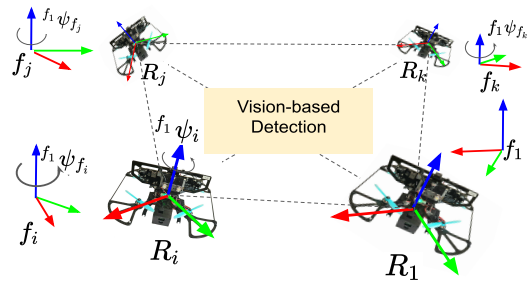}
    \caption{An example of a homogeneous 4-MAV system featuring our Falcon 250 platform~\cite{mohta2018fast} of which dynamics and measurement models are used in the simulation. Each robot is associated with a body frame $i$ and a fixed frame $f_i$. They can communicate and visually detect each other anonymously with some probability}
    \label{fig:frame_def}
    \vspace{-.1in}
\end{figure}

Methods relying on exchanging environment features are called map localization~\cite{indelman2014multi, montijano2016vision, wasik2019robust, guo2017ultra, vemprala2016vision}. For example, Montijano et al.~\cite{montijano2016vision} and Leahy et al.~\cite{leahy2019control} propose to use homography estimation~\cite{nguyen2018unsupervised} to compute the relative pose between two robots. The main problem with this approach is that the robots either need to maintain a map of features or have overlapping views with shared features, not to mention the challenge in finding good features in low or texture-less environments. 
Another methods, including this study, rely on direct robot-to-robot measurements, is called mutual localization in~\cite{franchi2009mutual}.  

Our study focus on localization for systems of multiple micro Aerial Vehicles (MAVs) which have relatively small sizes and weights than unmanned aerial vehicles and suitable for multiple applications such as surveillance, and search and rescue operations. Unlike ground vehicles, MAVs are subject to size, weight and payload (SWaP) constraints. 
As a result, cameras are often preferred over range finders and LiDARs to do the same task due to their compactness and light weight. The system of robots studied in this work, as shown in Fig.~\ref{fig:frame_def}, are equipped with two types of sensor modalities: vision-based detection and measurement of other robots within each robot's field of view; and visual inertial odometry (VIO) using stereo cameras and inertial measurement unit measurements. These robots are \SI{250}{\cm} in length, inexpensive and light-weight, making them a great candidate for research and civilian applications. 

Despite its appeals, mutual localization in such multi-MAV systems is challenging due to a couple of reasons. First, our system of robots is homogeneous, visually similar, and using tagging or specific sensors for identifying robots are neither practical or desirable due to the SWaP constraints. Thus, the vision-based detection provides no identify information, leading to data association ambiguity. 
Secondly, the vision-based detection often yields false positives and false negatives. Furthermore, unlike range finders used in previous studies such as~\cite{franchi2013mutual}, the vision-based measurements of distance and bearing are quite noisy. These factors make the data association problem become even more challenging. 

In short, we study the problem of multi-MAV mutual localization under following assumptions, 
\begin{itemize}
    \item Initial relative poses between robots are unknown.
    \item Robot detection provides no identity information.
    \item Robot detection can include false negatives and false positives.
    \item Vision-based distance and bearing measurements are noisy.
\end{itemize}

Our main contributions in this work include: 
\begin{itemize}
    \item We introduce a vision plus IMU framework for localization that uses relative distance and bearing measurements, on a SWaP-constraint platform.
    \item We propose an extension of CPDAF with a simple but effective gating and evaluating mechanism to keep the number of hypotheses manageable. 
    \item We demonstrate how on-board sensing, estimation and control can be used for formation flight. 
\end{itemize}

Our method is applicable to a system with any kind of distance and bearing sensors but the implementation introduced in this work focuses on using camera sensors for these measurements.
\section{Related Work}
\label{sec:related_work}
Mutual localization has attracted a large amount of research works. For example, Spica et al.~\cite{spica2016active} address the problem of estimating the formation scale in the context of bearing-based formation localization for multiple robots. In~\cite{mariottini2005vision}, authors propose an Extended Kalman Filter for the estimation of each follower position and orientation with respect to the leader, using bearing information only. However, those works do not consider the problem caused by unknown data association which plays an important role in mutual localization. For instance, Mehrez et al.~\cite{mehrez2017optimization} assume that robots are able to uniquely identify each of the observed robots in their field of view and measure their relative range and bearing. 

The literature has investigated in providing relative measurements with robot identities via tagging. Some recent examples using this method for relative localization of a team of aerial robots include~\cite{tron2016distributed} where authors use colored circular markers on the robots to obtain relative bearing between the robots. De silva et al.~\cite{de2014ultrasonic} develop 3D sensor nodes employing ultrasonic-based range measurement and infrared-based bearing measurement for spatial localization of robots. Dias et al.~\cite{dias2016board} utilize active markers to identify unique ID of quadrotors based on pulsating at a predefined frequency. The main disadvantage of these methods is that they do not scale well with the number of robots. 

Recent approaches directly deal with unknown robots' identities and attempt to estimate these identities together with robot localization. Chang et al.~\cite{chang2011vision} propose a maximum likelihood data association algorithm with a
threshold gating on the Mahalanobis distance between the
incoming measurement and the expected measurement. The problem with this method is that the selected measurement may not be the correct one due to the inaccuracy of the measurements, leading to filter divergence. In this work, we propose a probabilistic data association framework that can handle the noisy measurements. 

The problem of mutual localization with anonymous relative measurements was first considered by Franchi et al.~\cite{franchi2009mutual,franchi2010solvability, franchi2013mutual}. In~\cite{franchi2009mutual}, authors introduce a two-phase localization system in which a multiple registration algorithm to build data association hypotheses is followed by a Multi-Hypothesis EKF to prune out hypotheses inconsistent with the current belief. Their successive work~\cite{franchi2010solvability} proposes to feed back the belief in system's spatial configuration to handle the worst case scenario where the computation can be a factorial function of number of agents when the spatial arrangement of the robots is rotational symmetric. In~\cite{franchi2013mutual}, they improve the algorithm further by using particle filters to replace EFK filters. Compared to our work, theirs rely on an assumption that the posterior probability distribution functions of robot states are independent so that each particle filter can be feasibly updated in a separate manner. Their frameworks also suffer from adding computation to maintain and update particle filters,  especially when scaling up the number of robots or in case the robot state has a high dimension.

\section{Problem Formulation}
\label{sec:problem}
Let us consider a problem formulation with a team of homogeneous $N-$robots $\{\robot_1, \robot_2, \dots \robot_N\}$, $N$ is known. 
Beside the the attached moving frame $i$, each robot $\robot{i}$ is attached to a fixed frame $f_i$ as shown in Fig.~\ref{fig:frame_def} such that the $Z_i$ axis of the frame $f_i$ is on the same direction with the gravity. In the follows, we describe the mutual localization problem whose objective is to localize every $\robot{i}$ to a common frame, which can be any in the set $\{f_1, f_2, \dots f_N\}$. Without loosing the generalization, we choose $f_i$ to be $f_1$. 

Suppose $\trans{i}{f_j} \subset \mathbb{R}^3$ and $\rot{i}{f_j} \subset \mathcal{SO}3$, $i \in \{1, \dots N \}$ denote the translation and rotation between robot $\robot{i}$ and frame $f_j$, respectively. Then, localizing robot $\robot{i}$ in frame $f_1$ is equivalent to estimate $(\trans{i}{f_1}, \rot{i}{f_1})$. We can define a set $\estimatesystemstate{t} = \{ (\trans{i}{f_1}, \rot{i}{f_1})\ | i \in \{1, \dots N \} \}$ involving all variables that we aim to estimate. 
Our problem becomes, 
\begin{equation}
    \estimatesystemstate{t|t \in [0,T]} = \argmin_{\estimatesystemstate{t}}\ \estimatecost{t}{t}\ = \argmin_{\estimatesystemstate{t}}\ || \estimatesystemstate{t} - \truesystemstate{t} ||_2 ^2 
    \label{eq:cost}
\end{equation}
where $\truesystemstate{t}$ is the ground truth.

\section{The Stochastic Model}
\label{sec:stochastic_model}
Before representing the proposed approach, we first define discrete models for the system dynamics and observation measurements.   
\subsection{The System State Model}
Looking from the chosen common frame $f_1$, we have, 
\begin{equation}
    \begin{split}
        \trans{i}{f_1} & =\ \rot{f_i}{f_1}\ \trans{i}{f_i}\ +\ \trans{f_i}{f_1}\ \ \forall i \in {1, \dots, N} \\  
        \rot{i}{f_1} & =\ \rot{f_i}{f_1}\ \rot{i}{f_i} \hspace{.65in} \forall i \in {1, \dots, N} \\  
        \trans{j}{f_1} & =\ \rot{i}{f_1}\ \trans{j}{i}\ +\ \trans{i}{f_1} \hspace{.28in}  \forall i,j \in {1, \dots, N}
    \end{split}  
    \label{eq:common_frame}
\end{equation}

Thus, the odometry measurement comes from the VIO system of robot $\robot{i}$, 
\begin{equation}
    \measure{i}{} =  
    \begin{bmatrix} 
        \trans{i}{f_i} \\
        \rot{i}{f_i}
    \end{bmatrix}
    = 
    \begin{bmatrix} 
         \rot{f_i}{f_1}^T ( \trans{i}{f_1}\ -\  \trans{f_i}{f_1} ) \\
         \rot{f_i}{f_1}^T\ \rot{i}{f_1}
    \end{bmatrix}
    \label{eq:absolute_measurement}
\end{equation}

The detection measurement generating from robot $\robot{j}$ detected by robot $\robot{i}$ can be presented as, 
\begin{equation}
    \measure{j}{i} =  
    \begin{bmatrix} 
        \trans{j}{i}
    \end{bmatrix}
    = 
      \begin{bmatrix} 
        \rot{i}{f_1}^T\ ( \trans{j}{f_1}\ -\ \trans{i}{f_1} )
    \end{bmatrix}
    \label{eq:relative_measurement}
\end{equation}

    
The first two equations in Eq.~\ref{eq:common_frame} show that to achieve the rotation and translation of robot $\robot{i}$ in frame $f_1$, given only local measurements, we need to know the rotation and translation of frame $f_i$ in $f_1$. Thus, we define the state of robot $\robot{i}$ as  
$$\state{i} = \left[ \trans{i}{f_i}^T\ \rot{i}{f_1}\ \trans{f_i}{f_1}^T\ \rot{f_i}{f_1} \right]^T$$
Note that, we substitute $\trans{i}{f_1}$ by $\trans{i}{f_i}$ to make it convenient to define the state equation and that these two variables can be derived from each other. The coupled state system can be defined as 
$$\state{} = [\state{1}^T\ \state{2}^T\ \dots\ \state{i}^T \dots\ \state{N-1}^T \state{N}^T]$$

We can decompose the rotation $\rot{i}{f_1}$ into  two parts, $\rot{z}{}(\yaw{i}{f_1})$ corresponds to the rotation around the gravity vector, and $\rot{i,xy}{f_1}$ corresponds to the rotation on the  plane perpendicular to the gravity vector, 
\begin{equation*}
    \rot{i}{f_1} =\ \rot{z}{}(\yaw{i}{f_1}) \rot{i,xy}{f_1}
\end{equation*}
In a VIO system, only rotation along $Z$ axis is unobservable. Thus, we can assume that $ \rot{i,xy}{f_1}$ is known, leaving only $\yaw{i}{f_1}$ needs to estimate. Furthermore, every frame $f_i$ is defined to be different only in the rotation around $Z$ axis, denoted as $ \yaw{f_i}{f_1}$. Thus, 
   $\rot{f_i}{f_1} =\ \rot{z}{}(\yaw{f_i}{f_1})$
and we can rewrite the individual robot state, 
$$\state{i} = \left[ \trans{i}{f_i}^T\ \yaw{i}{f_1}\ \trans{f_i}{f_1}^T\ \yaw{f_i}{f_1} \right]^T$$



We utilize a linear system whose input is the velocity. The velocity input is assumed to be corrupted with i.i.d zero-mean Guassian noise. The robots' position and yaw can be modelled as follows 
\begin{gather*}
    \pdot{i}{f_i}   = \vtrans{i}{f_i}\  +\ \gnoise{v_i}{f_i} \ , \quad    
    \yawdot{i}{f_1} = \vyaw{i}{f_1}\  +\ \gnoise{\omega_i}{f_1} 
\end{gather*}

Frames $f_2, \dots f_N$ have unknown, fixed transformations  with respect to frame $f_1$ but there can be drifts due to errors from VIO systems, 
\begin{gather*}
    \pdot{f_i}{f_1}   =  0\  +\ \gnoise{p_i}{f_1} \ , \quad
    \yawdot{f_i}{f_1} =  0\  +\ \gnoise{\psi_i}{f_1} 
\end{gather*}

We can write the state equation in a standard form
\begin{equation*}
    \dotstate = A \state{} + B \cinput{} + \statenoise{}
\end{equation*}
where $ A = \mathbf{0}_{7N},\ \statenoise\ \sim \mathcal{N}(0, Q)$
\begin{equation*}
\vspace{0.1in}
    B = \begin{bmatrix}
            I_4&          \Bold{0}{4}& \Bold{0}{4}& \dots&  \Bold{0}{4} \\ 
            \Bold{0}{4}&  \Bold{0}{4}& \Bold{0}{4}& \dots&  \Bold{0}{4}  \\ 
            \vdots&       \vdots&      \vdots&      \ddots& \vdots     \\ 
            \Bold{0}{4}&  \Bold{0}{4}& \Bold{0}{4}& \dots&  I_4&       \\
            \Bold{0}{4}&  \Bold{0}{4}& \Bold{0}{4}& \dots&  \Bold{0}{4} 
        \end{bmatrix},\ 
    \cinput\ = 
        \begin{bmatrix}
            \vtrans{1}{f_1}\\
            \vyaw{1}{f_1} \\ 
            \vdots \\
            \vtrans{N}{f_N} \\ 
            \vyaw{N}{f_1}
        \end{bmatrix}
\end{equation*}
$Q$ is the covariance matrix of the i.i.d Gaussian noise.
We discretize this continuous time system using zero-hold for the input 
\begin{equation}
    \state{k} = F\state{k-1} + G \cinput{k-1} + \statenoise{k-1} 
    \label{eq:system_model}
\end{equation}
where $k=1,\dots, T$ is the current time step, $F = I_{7N}, G = B \Delta t$, $\statenoise{k-1} \sim \mathcal{N}(0, Q_d)$, $Q_d = Q \Delta t$, $\Delta t$ is the sampling time. 

\subsection{The Measurement Model}
During the update process, we update two types of measurements in a decoupled manner.  
\subsubsection{Odometry Measurements} 
Odometry measurements from VIO systems directly provide each robot's individual state with respect to its own frame. We rewrite Eq.~\ref{eq:absolute_measurement} as follows, taking into account a Gaussian noise,   
\begin{equation*}
    \measure{i}{}\ = 
    \begin{bmatrix}
        \trans{i}{f_i}\ +\ \mnoise{\trans{i}{}}{f_i}  \\
         \rot{z}{}(\yaw{f_i}{f_1} + \mnoise{\yaw{f_i}{}}{f_1} )^T\
        \rot{z}{}(\yaw{i}{f_1} + 
     \mnoise{\yaw{i}{}}{f_i})\ \rot{i,xy}{f_1}
         
    \end{bmatrix}
\end{equation*}
where $\mnoise{(.)}{(.)}$ denotes noises. 
There is no data association involved in this partial update step. 

\subsubsection{Detection Measurements}
Detection measurements from the vision-based detection can help estimate the pose of robots detected. We assume that robots can detect all other robots with detection probability $P_D$ and some probability of false positive, false negative. This assumption can be achieved using $360^o$-cameras and that robots are in the range of detection. 
Eq.~\ref{eq:relative_measurement} can be rewritten with white noise added, for true measurements,
\begin{equation*}
    \measure{j}{i} =\  \rot{z}{}(\yaw{i}{f_1})^T\ (\trans{j}{f_1}\ -\ \trans{i}{f_1})\ +\ \gnoise{j}{i}
\end{equation*} 
Let $F_{t=0,\dots,K}$ be random variable representing the number of false positives at time $t$. We assume $F$ to has Poisson distribution, 
\begin{equation*}
    P_{F_t}(F) = \exp{(-\lambda V)} (\lambda V)^F / (F!)
\end{equation*} 
\subsubsection{Detection Measurement Permutation}
To handle the unknown data association as well as the false positives, false negatives, we define the following helper variables, similar to those in~\cite{cpdaf}. Note that for each robot, we have $N-1$ targets.  
\begin{itemize}[leftmargin=*]
    \item $M_i$, number of measurements at current time on robot $\robot{i}$
    \item $\phi_{i,j} \in {0,1}$, an indicator that tells whether robot $\robot{j}$ is detected by robot $\robot{i}$ among $M_i$ target measurements  
    \item $\detector{i}$, a $(N-1)$-dimensional vector stack of all $\phi_{i,j}$ 
    \item $D_i= \sum_{i=1}^N\ \phi_{i,j}$, the number of detected robots on robot $\robot{i}$
    \item $\pertubmatrix{i}$, a $D_i \times M_i$, a permutation of $D_i$ true measurements among $M_i$ relative measurements at current time
\end{itemize}
Given $M_i$ measurements on robot $\robot{i}$, it can be understood that $\detector{i}$ is a possible outcome on which robots are actually detected among $(N-1)$ targets. There can be $2^{N-1}$ such $\detector{i}$ outcomes. Given an outcome $\detector{i}$, there could be many possible ways to match detection measurements with the detected robots. We do not know for sure since there is no identity information in the detection measurements. Each $\pertubmatrix{i}$ is a possible match. Thus, by combining $\detector{i}$ and $\pertubmatrix{i}$, we can cover all possible association events happened to the $M_i$ detection measurements on robot $\robot{i}$. We call a tuple  $(\detector{i}, \pertubmatrix{i})$ a data association hypothesis, or simply a hypothesis. The approach introduced in this work centers around finding all feasible $(\detector{i}, \pertubmatrix{i})$-hypotheses and updating the system state based on the probability of each hypothesis, for every robot $\robot{i}$.

\section{Method} 
\label{sec:method}

As can be seen from Eq.~\ref{eq:relative_measurement}, each detection measurement depends on the state of multiple robots, making it improper to use the standard JPDA filter~\cite{bar2009probabilistic} for state estimation. Instead, we develop an extension of CPDA filter~\cite{cpdaf} for the nonlinear measurement model to update the system state over time using both detection measurements and odometry measurements. 
Our approach iteratively treats each robot as a station while others are targets and applies CPDAF on the system. To simplify the notations, in the follows, we represent our extension of CPDAF in case a robot $\robot{}$ served as the station with $L$ targets, $M$ detection measurements and $1$-D dimensional states.

\subsection{CPDAF Extension for Nonlinear Measurement Model}
\subsubsection{CPDAF Step 1 - Prediction}
We denote $\state{k-1|k-1}, \statecov{k-1|k-1}$ as the state and covariance of the state at time step $k-1$, respectively. A prior estimate for the state and covariance at time step $k$ is obtained by probagating through the system model in Eq.~\ref{eq:system_model}, 
\begin{align}
    \state{k|k-1} &= F \state{k-1|k-1} + G \cinput{k-1}  \\
    \statecov{k|k-1} &= F \statecov{k-1|k-1} F^T + Q_d \nonumber
    \label{eq:cpdaf_predict}
\end{align}

\subsubsection{CPDAF Step 2 - Gating}
\begin{figure}
   \begin{minipage}{.48\linewidth}
      \includegraphics[width=\linewidth]{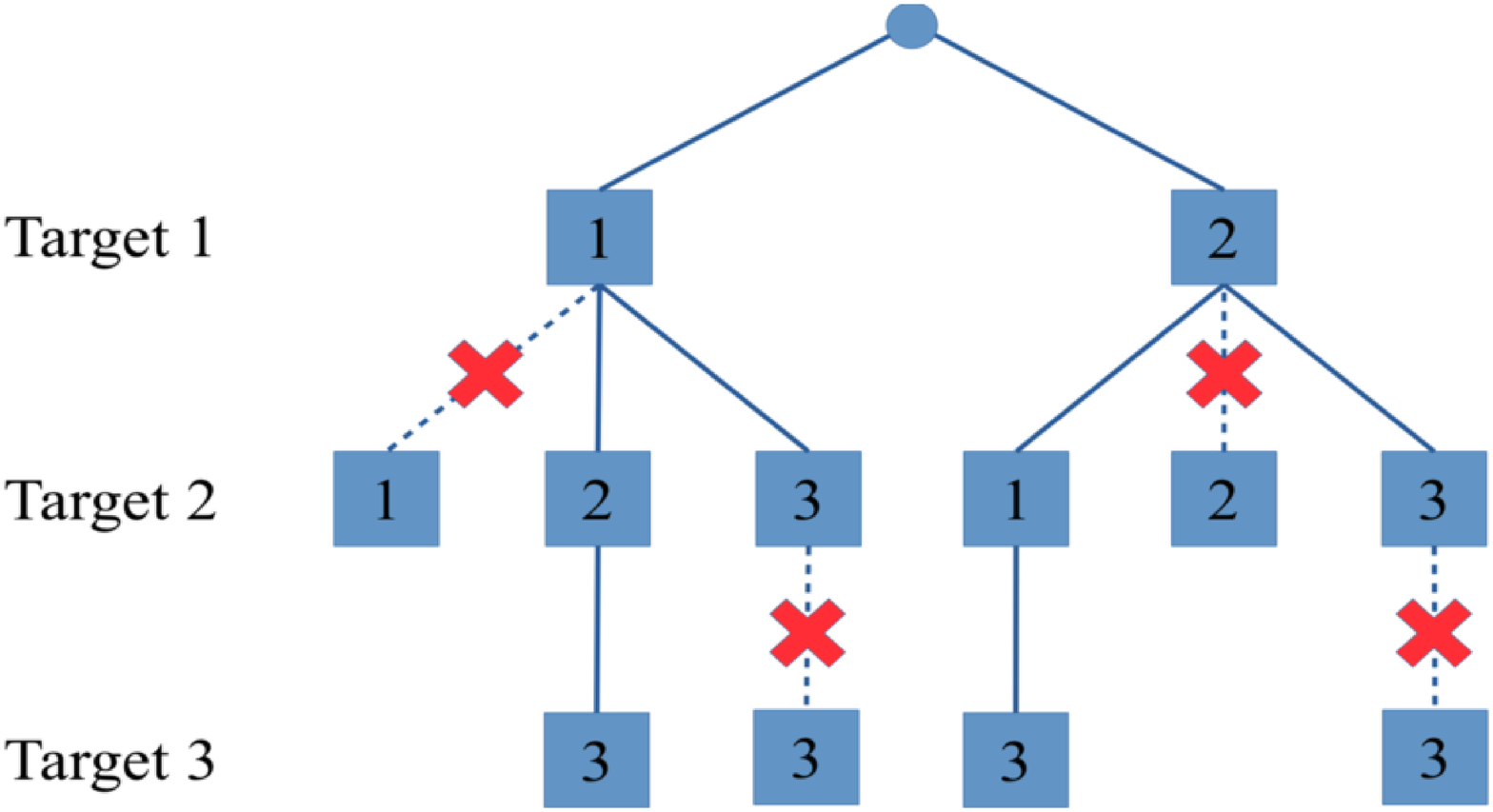}
      \vspace{0.1in}
      \subcaption{}
    \end{minipage}
    \hfill
    \begin{minipage}{.46\linewidth}
       \includegraphics[width=\linewidth]{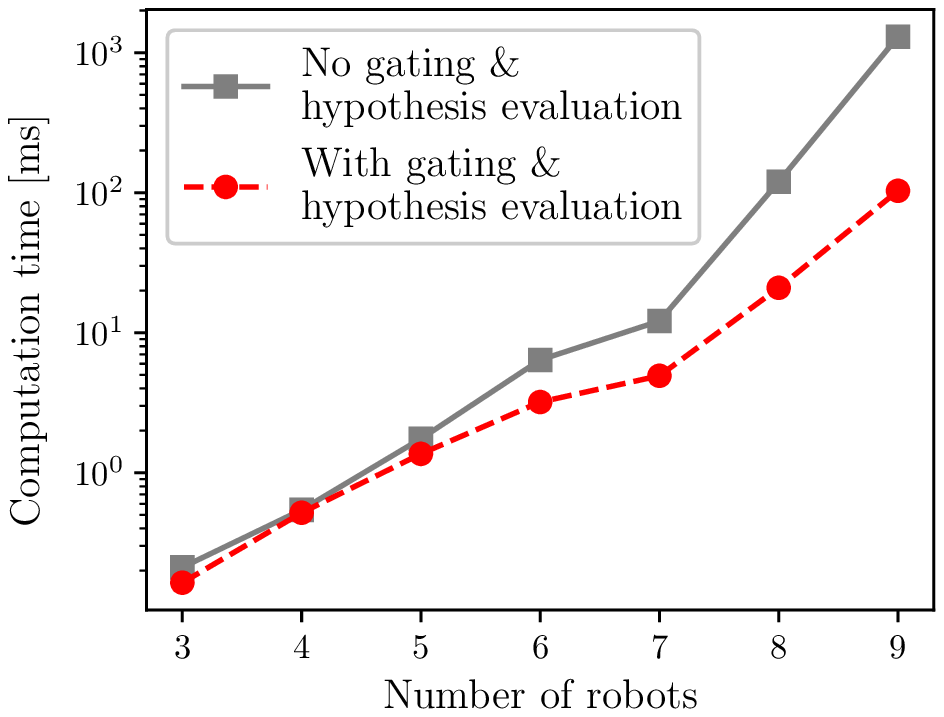}
       \subcaption{}
    \end{minipage} 
    \caption{ 
    (a): Hypothesis tree for three targets and three
 measurements after gating. $\emptyset$ is omitted for simplicity. The set of valid hypotheses is $\{ (1,2,3), (2,1,3) \}$\\
    (b): Comparison of the computation time required for the CPDAF update step with and without gating and hypothesis evaluation. Note that the Y-axis has log scale
    }
    \label{fig:hypothesis_tree}
    \vspace{-.1in}
\end{figure} 

This step aims to reduce the number of possible measurements in the $M$-measurement set that can be assigned to each target robot. In our problem, the detection measurement is nonlinear, due to Eq.~\ref{eq:relative_measurement}. Thus the gating for target robot $j$
\begin{equation}
    G_j \defeq \{ z_j | (z_j - h_j(\state{})) ^T P_{z_j z_j}^{-1} (z_j - h_j(\state{})) \leq \gamma\}
\end{equation}
Where $P_{z z}$ is the innovation covariance matrix computed as same as in UKF~\cite{wan2000unscented} and $P_{z_j z_j}$ is the block of this matrix corresponding to target $j$.  $\gamma$ is a threshold taken from the inverse chi-squared cumulative distribution at a significance level $P_G$ and the degree of freedom equal to dimension of $h_j( \state{})$.  

\subsubsection{CPDAF Step 3 - Evaluation of Hypotheses}
As~\cite{crouse2010jpdaf} pointed out, the total number of $(\detector{}, \pertubmatrix{})$-hypotheses for a set of measurement $M$ on a robot $\robot{}$ is
\begin{equation}
    \sum_{D}^{\min(M, L)} 
    \binom L{D}
    \binom {L}{D}
    D!
\end{equation}
This can make evaluating them over time intractable when $L$ and $M$ are large. 
However, this assumption is not valid in our case where a detection measurement  depends on the state of two robots. To tackle this problem, we propose an efficient evaluating algorithm.  
This algorithm, inspired by~\cite{zhou1993multitarget}, starts by creating an association hypothesis tree of depth $L$ where each level of the tree represents the matching for a target robot. Each level consists of some nodes representing all possible detection measurements that can be assigned to that target, including $\emptyset$ - an indication that the target is not detected. Thus, a valid hypothesis is a path connecting the root with one single node on every level such that each node exists exactly one time on the path, except $\emptyset$. The remaining of this algorithm is to do a depth-first traversal to find all those paths. We omitted the algorithm's details here due to the space limit.
An illustration of the algorithm is shown in Fig.~\ref{fig:hypothesis_tree}(a). 
 

\subsubsection{CPDAF Step 4 - Measurement-based Update}
As mentioned in section~\ref{sec:stochastic_model}, the measurement-based update is separated into two update steps. The first update is based on the odometry measurement, as same as in UKF~\cite{wan2000unscented}, and the second update is based on the detection measurement.  

The later update is based upon the list of valid hypotheses obtained after step 2 and step 3. We extend CPDAF to compute the probability of a hypothesis $(\detector{}, \pertubmatrix{})$~\cite{cpdaf} in case the measurements are nonlinear,  
\begin{align*}
  \beta(\detector{}, \pertubmatrix{}) =  \frac{1}{c} F(\detector{}, \pertubmatrix{}) . \lambda^{L - D}\
  \prod_{i=1}^L   (1 - P_D) ^ {1- \detector{i}} (P_D)^{\detector{i}}  
\end{align*}
where $c$ is the normalization factor, $\lambda$ is the false observation spatial density, $P_D$ is the detection probability of a target robot. $P_D$ can vary among robots. 
\begin{equation*}
    F(\detector{}, \pertubmatrix{}) = \frac{\exp{\left( \mu(\detector{}, \pertubmatrix{})^T S(\detector{})^{-1} \mu(\detector{}, \pertubmatrix{}) \right)}} 
    {\sqrt{(2\pi)^{D} \det(S(\detector{}))}}
\end{equation*}
where
\begin{align*}
    \mu(\detector{}, \pertubmatrix{}) &= \pertubmatrix{} \mathbf{z} - \Phi(\detector{}) \mathbf{h}(\state{}) \\
    S(\detector{}) &= \Phi(\detector{}) P_{zz} \Phi(\detector{}) ^T
\end{align*}
with $\state{}$ is the system state at current time which is $(L+1)$ dimensional, $H$ is the $L \times (L + 1)$ dimensional observation matrix, $\mathbf{z}$ is the $M$-dimensional vector of stacked detection observations, $\Phi(\detector{})$ is a $D \times N$ binary matrix with $r^{th}$ row equal to $r^{th}$ non-zero row of diag$(\detector{})$, and $P_{zz}$ is the innovation covariance matrix, computed as same as in UKF~\cite{wan2000unscented}. 


Based on~\cite{cpdaf}, we derive the state and covariance udpate as follows. 
\begin{align*}
    \state{k|k} &= \state{k|k-1} + \sum_{ \detector{}} K(\detector{}) \sum_{ \pertubmatrix{}} \beta(\detector{}, \pertubmatrix{}) \mu(\detector{}, \pertubmatrix{}) \\
    \statecov{k|k} &= \statecov{k|k-1} -  \sum_{  \detector{}} K(\detector{}) \phi(\detector{i}) P_{zx}\sum_{\ \pertubmatrix{}} \beta(\detector{i}, \pertubmatrix{}) \\
\noindent    & + \sum_{\detector{}} K(\detector{})  \left( 
                   \sum_{  \pertubmatrix{}} \beta(\detector{}, \pertubmatrix{}) \mu(\detector{}, \pertubmatrix{}), \mu(\detector{}, \pertubmatrix{})^T \right)       K(\detector{})^T \\ 
    & - \left( \sum_{ \detector{}} K(\detector{})  
                  \sum_{  \pertubmatrix{}} \beta(\detector{}, \pertubmatrix{}) \mu(\detector{}  
       \right)  \\
   & \times \left( 
    \sum_{ \detector{}} K(\detector{}')  
                  \sum_{  \pertubmatrix{}'} \beta(\detector{}', \pertubmatrix{}') \mu(\detector{}' \right)
\end{align*}
where 
$K(\detector{}) = (\Phi(\detector{}) P_{zx})^T P_{zz}^{-1}$
is the Kalman gain, $P_{zx}$ is the measurement-state cross-covariance  matrix.

\subsection{Time Complexity Analysis}
Step 1 - executing model prediction takes $\mathcal{O}(N)$. 

Step 2 iterates over $M$ target measurements on every robot $\robot{}$ to prune out measurements that are not in the validation gate. Step 2 has $\mathcal{O}(\sum_{i=1}^N M) = \mathcal{O}(MN)$ time complexity.

Step 3 essentially carries out a depth-first traversal over all nodes and edges. 
In the worst case, each level, excepting the root, consists of $M + 1$ nodes and since every node in a level is connected to every node in the next level, there will be $\mathcal{O}(M^2)$ edges connecting two levels. Thus, the total of nodes and edges, or the worst case time complexity for traversing the tree is $\mathcal{O}(N M + N M^2) = \mathcal{O}(NM^2)$. 


The running time for the first update in step 4 is as same as in UKF~\cite{wan2000unscented}. The second update depends on how many valid hypotheses selected. In the worst case, this number is exponential with respect to the number of robots and measurements, making step 4 exponential in time of computation. In practice, this number is largely reduced thanks to step 2 and step 3. In addition, some technique such as $k$-best hypothesis can be used to make step 4 manageable. $k$-best hypothesis algorithm~\cite{murty, miller_optimize_murty} is $\mathcal{O}(kN^3)$. In our case, the complexity of finding $k$-best hypotheses on a tree is $\mathcal{O}(k\ \max(N,M)^3)$


\section{Experiments}
\label{sec:experiments}
\begin{figure}
   \begin{minipage}{.48\linewidth}
      \includegraphics[width=\linewidth]{{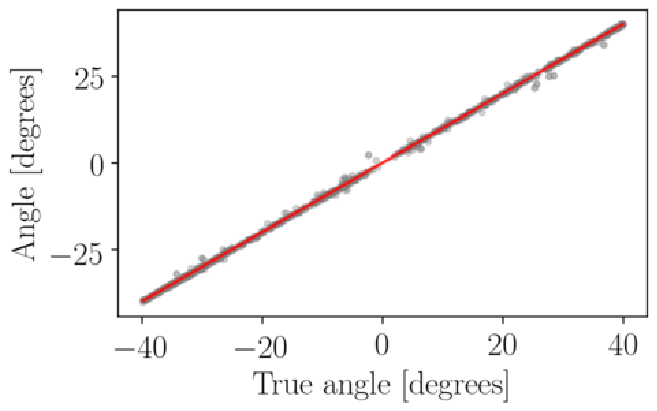}}
    \end{minipage}
    \begin{minipage}{.48\linewidth}
       \includegraphics[width=\linewidth]{{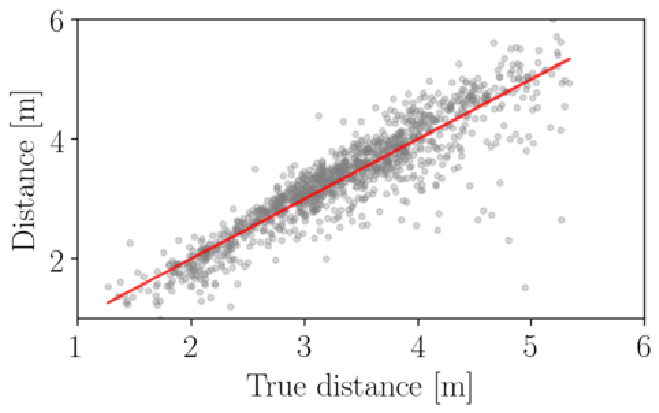}}
    \end{minipage} 
    \caption{Vision-based measurement models: (a) bearing; (b) distance. Red line: model v.s. true value, gray dots: measurement v.s. true value}
    \label{fig:detection_measurements}
    \vspace{-.2in}
\end{figure}

\begin{figure}
   \begin{minipage}{.48\linewidth}
      \includegraphics[width=\linewidth]{{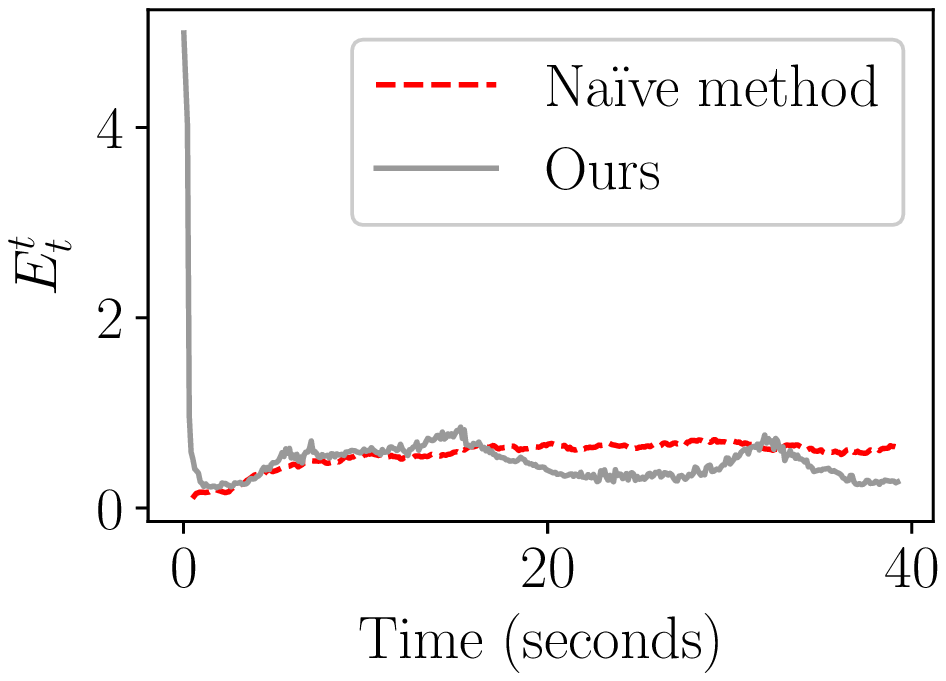}}
    \end{minipage}
    \begin{minipage}{.48\linewidth}
       \includegraphics[width=\linewidth]{{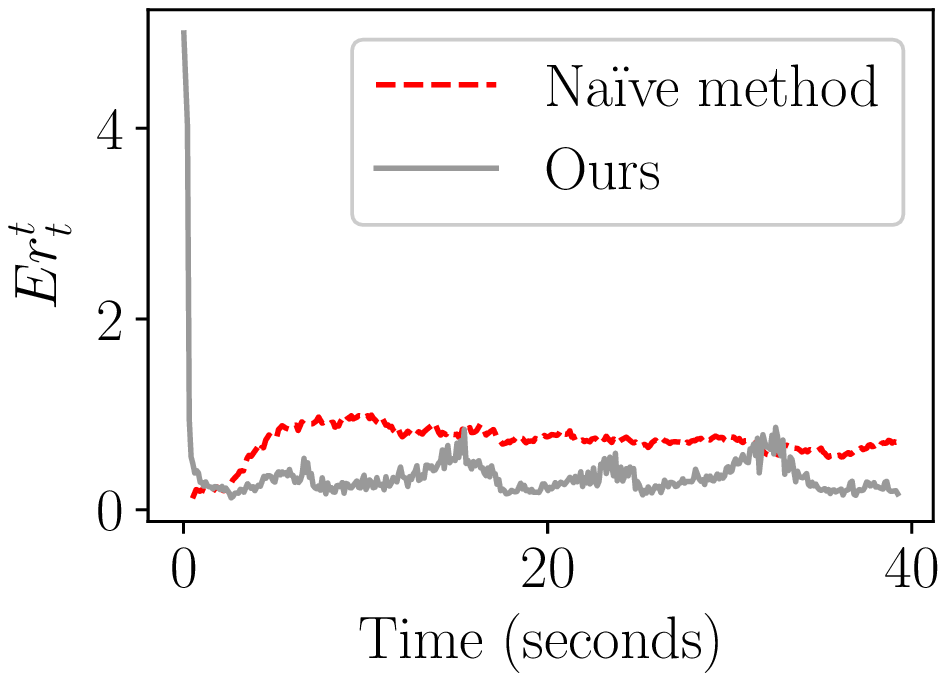}}
    \end{minipage} 
    \caption{
    From left to right: absolute and relative state errors during changes from \SI{1.35}{\m}-radius circle $\rightarrow$ \SI{2.7}{\m}-radius circle   $\rightarrow$  \SI{1.35}{\m}-radius circle  $\rightarrow$ \SI{2.7}{\m}-radius circle   $\rightarrow$  \SI{1.35}{\m}-radius circle}
    \label{fig:run_1}
    \vspace{-.1in}
\end{figure}

\begin{figure*}[h!]
  \centering
    \begin{minipage}{.245\textwidth}
      \includegraphics[width=\linewidth]{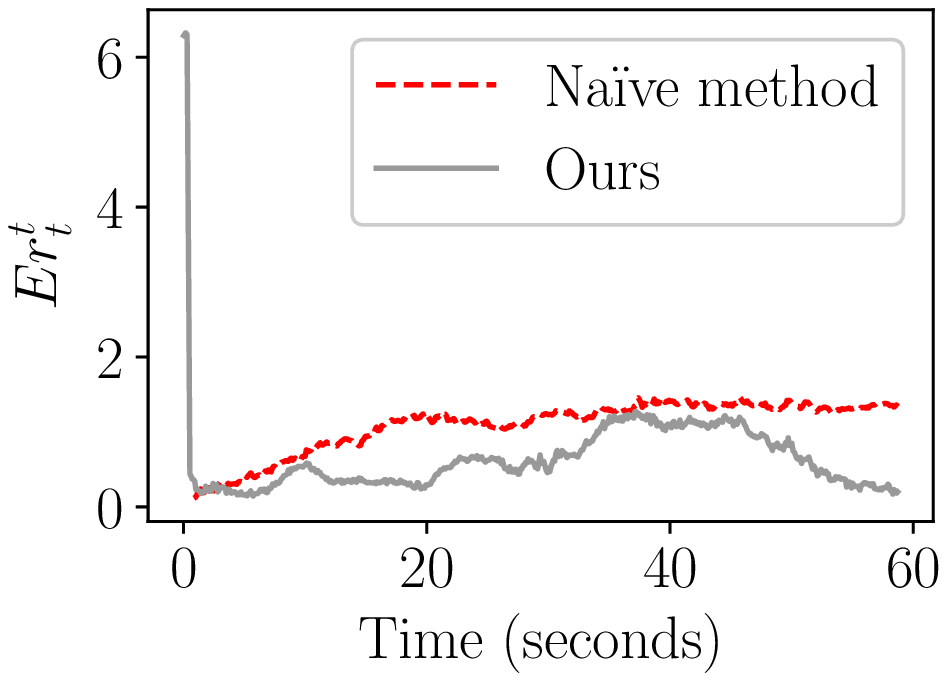}
      \subcaption{}
    \end{minipage}
    \begin{minipage}{.245\textwidth}
      \includegraphics[width=\linewidth]{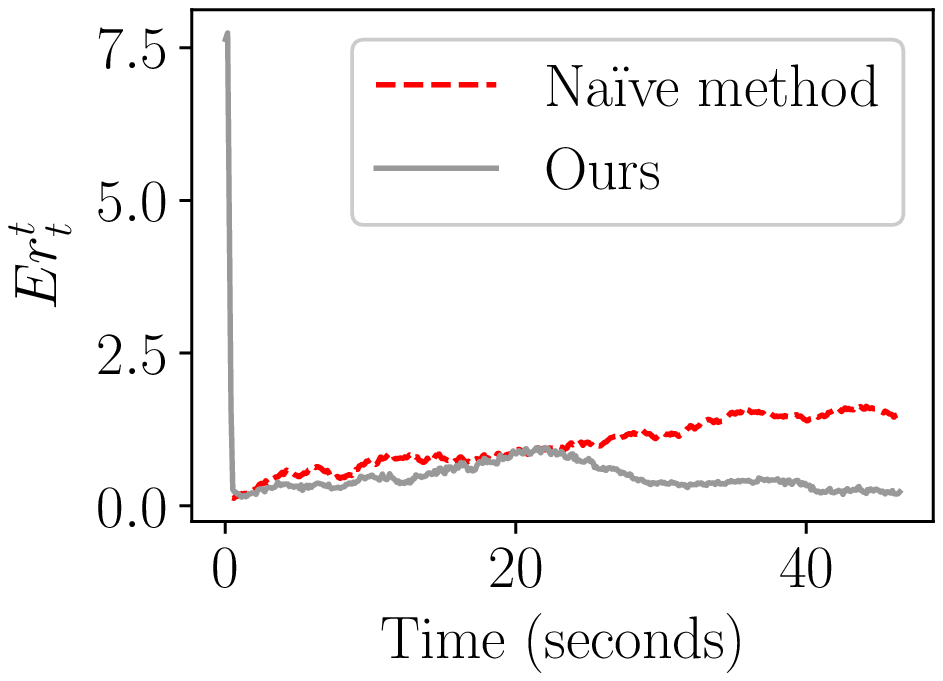}
      \subcaption{}
    \end{minipage}
    \begin{minipage}{.245\textwidth}
      \includegraphics[width=\linewidth]{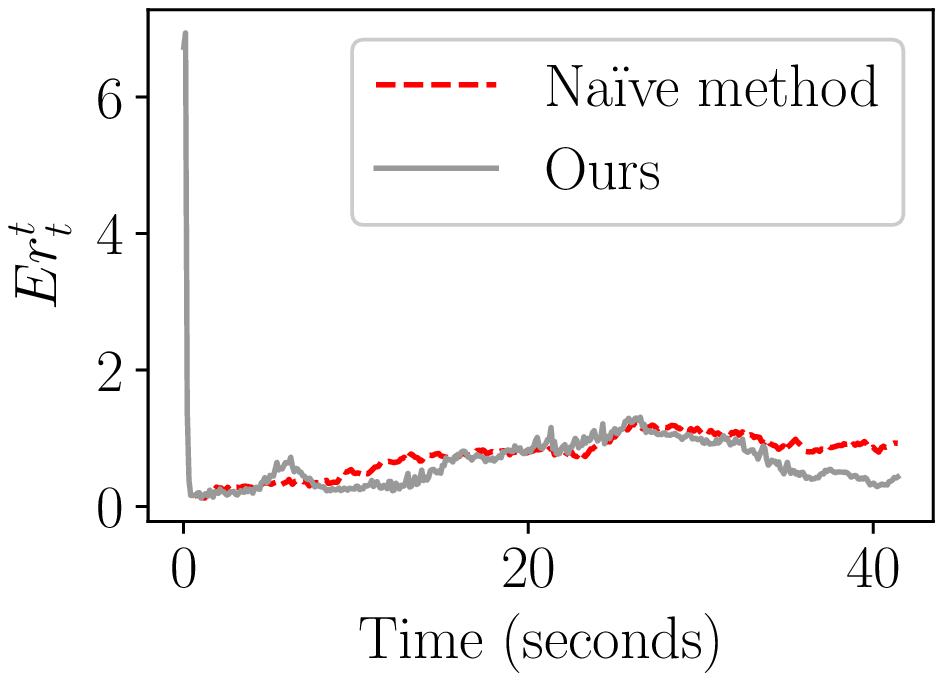}
      \subcaption{}
    \end{minipage} 
    \begin{minipage}{.245\textwidth}
      \includegraphics[width=\linewidth]{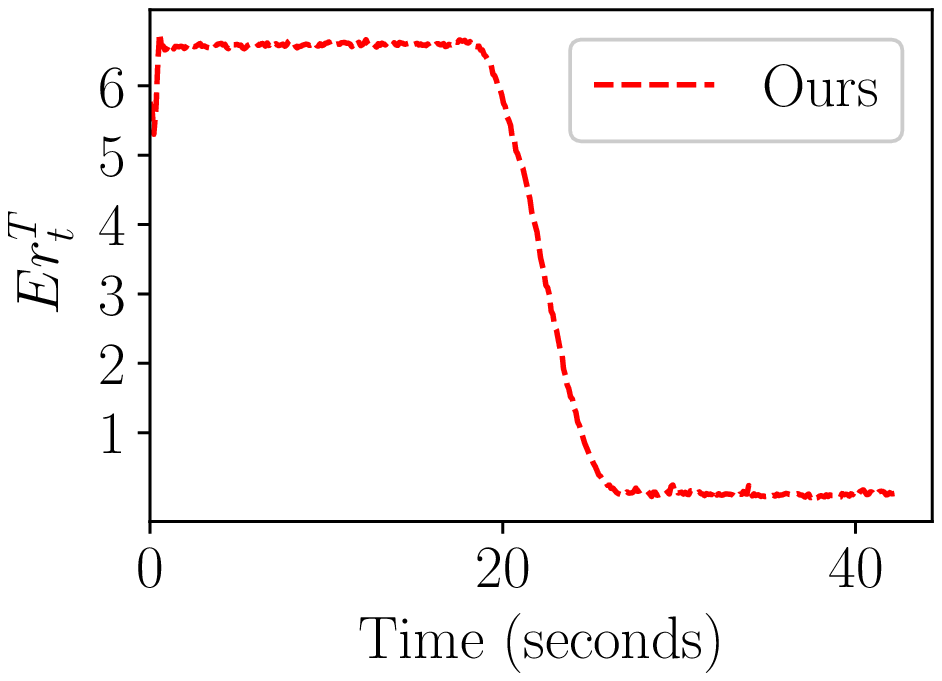}
      \subcaption{}
    \end{minipage} 
    
  \caption{(a,b,c): Relative state errors during the system's configuration changes. (a): \SI{1.35}{\m}-radius circle $\rightarrow$ v-line $\rightarrow$ line of \SI{12}{\m} $\rightarrow$ \SI{1.35}{\m}-radius circle $\rightarrow$ line of \SI{6}{\m}; (b): line of \SI{6}{\m} $\rightarrow$ line of \SI{12}{\m} $\rightarrow$ v-line $\rightarrow$ line of \SI{6}{\m}; (c): line of \SI{6}{\m} $\rightarrow$ \SI{1.35}{\m}-radius circle $\rightarrow$ \SI{2.7}{\m}-radius circle $\rightarrow$ line of \SI{12}{\m} $\rightarrow$ line of \SI{6}{\m}\\
  (d): Convergence of relative state to the desirable relative state as the robots take off and form a line of \SI{6}{\m}
  }\label{fig:configuration_change_performance}
  \vspace{-.1in}
\end{figure*}

We evaluate the efficiency of the proposed framework and algorithms on a simulation derived from experimentally-obtained measurement models.
\subsection{Simulation Settings}
The simulation is in ROS where the robots simulate our FLA Falcon 250 platforms as shown in Fig.~\ref{fig:frame_def}. In reality, each robot is featured with an Open Vision Computer~\cite{quigley2018open} consisting two gray-scale Python-1300 cameras to provide odometry measurements as well as detection measurements. Robots are simulated with \SI{0.54}{\m} in diameter and weigh \SI{0.5}{\kg}. The measurements that robots receive simulate measurement models that we obtain by doing real experiments on real robot platforms. 

\subsubsection{Odometry Measurement Model}
The VIO error is modeled as a multivariate Gaussian distribution,
$$\measure{i}{}|_{measured} = \measure{i}{}|_{true} + \mathcal{N}(0, \sigma_o^2)$$
where the standard deviation $\sigma_o$ is \SI{0.01}{\m} for elements in the transition and \SI{0.002}{\ rad} for euler angle elements in the orientation. 

\subsubsection{Detection Model}
We utilize MAVNet~\cite{MAVNet_tynguyen}, a light-weight and fast network for vision-based robot detection. The output segmentation is used to estimate the distance from the camera to the target as well as the bearing, assuming the robot's dimension in 3D world is known. 


The bearing measurement, as shown in Fig.~\ref{fig:detection_measurements}(a), is modelled as, 
$$bearing|_{measured} = bearing|_{true} + \mathcal{N}(0, \sigma_b^2)$$ 
where $\sigma_b =$ \SI{0.008}{\ rad}. 

The distance measurement as shown in Fig.~\ref{fig:detection_measurements}(b), is modelled as, 
$$distance|_{measure} = distance|_{true} + \mathcal{N}(0, \sigma_d^2)$$ 
with $\sigma_d = 0.0495*distance|_{true} + 0.0336$ \SI{}{m}. 

\subsection{Evaluation Metrics}
We use two metrics: 1) $\estimatecost{t}{t}$ in Eq.~\ref{eq:cost} that evaluates the \textit{absolute state} - the state of robots within the common fixed frame, $f_1$, and 2) $\relestimatecost{t}{t}$ that evaluates the \textit{relative state} - the state of robots within the common moving frame, i.e. frame $1$ of robot $\robot{1}$. 
\begin{equation}
    \relestimatecost{t}{t}\ =  || \estimaterelsystemstate{t} - \truerelsystemstate{t} ||_2 ^2 
    \label{eq:rel_cost}
\end{equation}
where $\estimaterelsystemstate{t} = \{ (\trans{i}{1}, \rot{i}{1})\ | i \in \{1, \dots N \} \}$ is the set of relative poses, $\truerelsystemstate{t}$ is the ground truth. 





\subsection{State Estimation Performance}
We evaluate the performance of our algorithm in comparison with a na$\Ddot{i}$ve method which assumes to know the initial system's configuration and integrates the VIO measurements over time.
In each experiment, a team of homogeneous $7$ robots are controlled to form different spatial configurations in a centralized manner. Each robot is controlled to follow a predefined path using its true state. 
The first experiment results depicted in Fig.~\ref{fig:run_1}, show that our state estimation converges very fast and matches the na$\Ddot{i}$ve method in the absolute state error while performs better in the relative state error. 
Fig.~\ref{fig:configuration_change_performance} also shows our method's superior performance in other three experiments in relative state errors. Absolute state errors are omitted since both methods perform similarly.

\subsection{Formation Control Using Estimated State}
We demonstrate an use case of our approach by designing an open-loop controller for formation flight. Our controller takes as input the initial estimate of the system state and determines an optimal path for every robot using minimum snap trajectory generation~\cite{mellinger2011minimum}. Each robot is then controlled to the final state using its current estimated state as the feedback. All steps, from state estimation to control is done on-board. 

We launch $7$ robots forming a line of \SI{6}{m} and evaluate $\relestimatecost{t}{T}$ which measures the convergence of the relative state estimate versus the final \textit{relative state}. 
\begin{equation}
    \relestimatecost{t}{T}\ =  || \estimaterelsystemstate{t} - \truerelsystemstate{T} ||_2 ^2 
    \label{eq:rel_cost_final}
\end{equation}
Fig.~\ref{fig:configuration_change_performance}(d) shows the error converges to $0$ illustrating that our controller is able to converge to the desire state when the robot team forms the desirable line. 

\subsection{Effectiveness of Gating and Hypothesis Evaluation}

  
Fig.~\ref{fig:hypothesis_tree}(b) shows the computation time of our proposed approach with various number of robots in two different settings: with and without using gating and hypothesis evaluation steps. Since these two steps eliminate unnecessary hypotheses, they keeps the algorithm run much faster. The more number of robots, the higher time saving can be achieved, starting from $1.29$ times with $3$ robots to $12.6$ times for $9$ robots. 

\section{Conclusions}
\label{sec:conclusion}
This work introduces CPDAF algorithm to address localization in SWaP-constrainted aerial platforms. We address three main challenges: 1) unknown data association in vision-based relative measurements and robot targets, 2) the need to boostrap the system from an unknown initial configuration, and 3) noisy vision-based measurements with false negatives, false positives. Experiments in simulation based on experimentally-derived models of measurements demonstrate the superior performance of our approach. We show how our state estimation can be used in a simple open-loop controller, extending the capability of using on-board sensing for estimation and control in formation flight. Our future work is to develop a close-loop controller as well as improve the hypothesis evaluation step. 
\bibliographystyle{IEEEtran}
\bibliography{root}

\end{document}